\documentclass{article}
\usepackage[final]{neurips_2023}
\makeatletter
\renewcommand{\@noticestring}{}
\makeatother
\usepackage[utf8]{inputenc}
\usepackage[T1]{fontenc}
\usepackage{hyperref}
\usepackage{url}
\usepackage{booktabs}
\usepackage{amsfonts}
\usepackage{amsmath}
\usepackage{nicefrac}
\usepackage{microtype}
\usepackage{xcolor}
\usepackage{graphicx}
\usepackage{algorithm}
\usepackage{algorithmic}
\usepackage{enumitem}
\usepackage{tikz}
\usetikzlibrary{arrows.meta, positioning, shapes.geometric, calc, decorations.pathreplacing}
\usepackage{multirow}
\usepackage{pifont}
\newcommand{\cmark}{\ding{51}}
\newcommand{\xmark}{\ding{55}}

\title{Memoir: Should a Model Write to Its Memory While It Thinks?}

\author{
  Jaber Jaber\thanks{Correspondence: \texttt{jaber@rightnowai.co}} \\
  RightNow AI\\
  \texttt{jaber@rightnowai.co} \\
  \And
  Osama Jaber \\
  RightNow AI\\
  \texttt{osama@rightnowai.co} \\
}

\begin{document}
\maketitle
\begin{center}
\includegraphics[height=1.1cm]{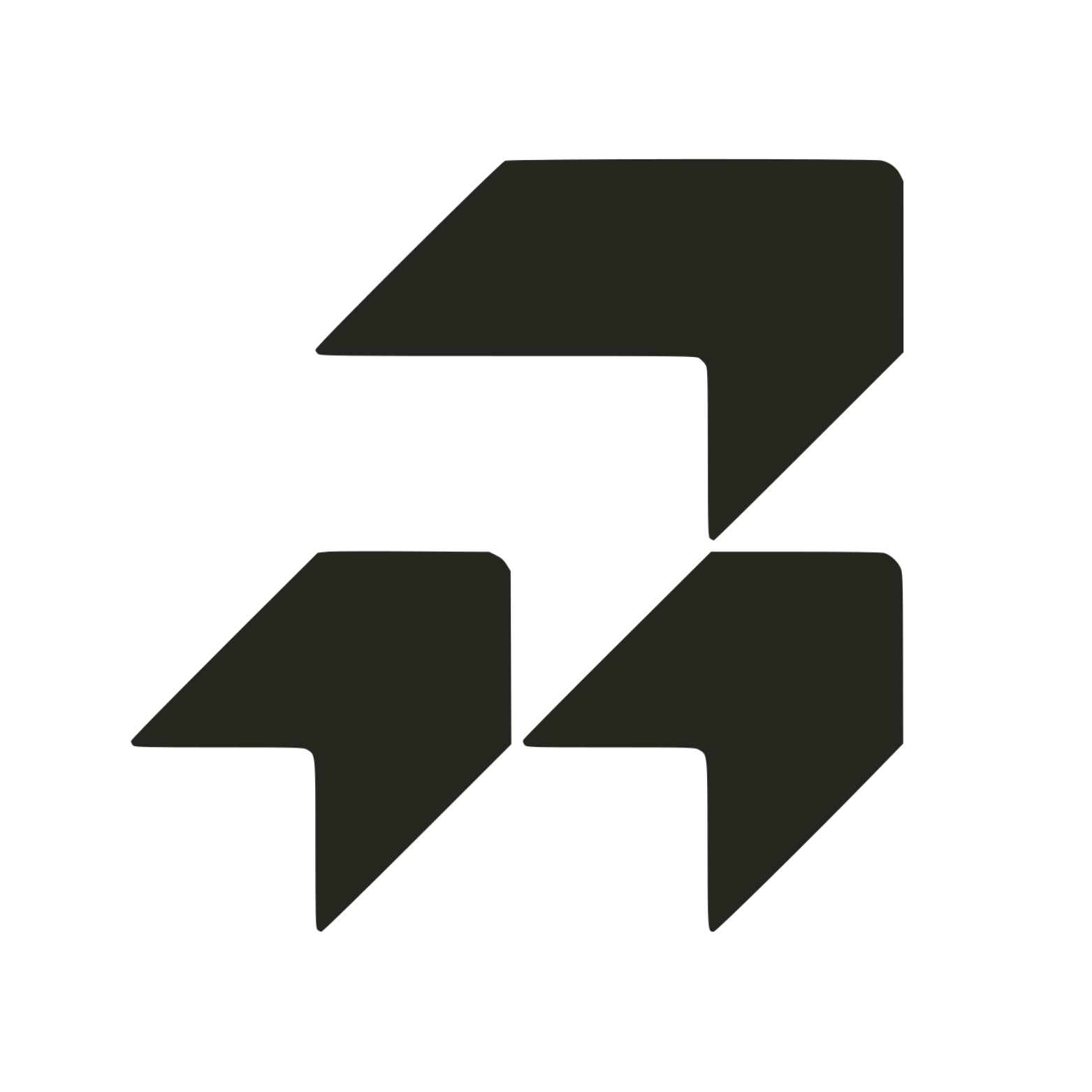}
\end{center}
\vspace{-0.3em}

\begin{abstract}
Memoir combines per-sample fast memory, shared slow parameters, variable-depth
latent recurrence, and a future-latent energy objective. We test its riskiest
coupling: each pondering iteration may rewrite the fast tier that the same
iteration reads. On procedural associative recall with key interference, we
compare a coupled arm against an otherwise identical read-only pondering arm.
Both arms contain 81,738 parameters, including 76,362 trainable parameters,
and use matched declared forward multiply-accumulate counts, data, optimizer,
schedule, and seeds. After 240 training steps across 12 seeds, coupled recall
is 0.5203 with a 95 percent interval of [0.4522, 0.5883], while read-only
recall is 0.6557 with [0.5953, 0.7160]. The arms are paired per seed, and the
read-only lead of 0.1354 gives a paired $t$ of 3.23 on 11 degrees of freedom
with a 95 percent interval of [0.0431, 0.2277] on the difference, winning on 10
of 12 seeds.
After 960 steps across 8 seeds, both arms reach 1.0000, so the measured effect
is a learning-speed penalty at a fixed budget, not a demonstrated capability
penalty. That longer control is ceiling limited, leaving convergence on a
non-saturating task unmeasured. A predicted failure in which memory rewriting
corrupts the energy signal did not occur: the energy margin grew and held.
Kernel restructuring also reduced delta-rule forward time from 0.907 ms to
0.351 ms on the stated device. Code and evidence are available at
\url{https://github.com/RightNow-AI/Memoir}
\end{abstract}

\section{Introduction}

Language models usually keep trained parameters fixed during a forward pass.
Long contexts and external retrieval expand what a model can read, but they do
not make recent experience part of the model's evolving weight state. Work on
test-time training and memorization instead treats the hidden state or a memory
module as an online learner \cite{sun2024ttt,behrouz2025titans}. Work on
adaptive recurrence lets a model spend more computation on difficult inputs
\cite{graves2016act,dehghani2018universal}. Memoir joins these directions in a
single interface: a recurrent latent process reads a hierarchy of weight tiers,
writes selected tiers, predicts future latents, and halts adaptively.

This integration creates a direct technical risk. If each pondering iteration
rewrites fast memory, then the recurrent operator changes while the latent is
being refined. An energy head could appear to improve because memory moved to
agree with the current latent, not because the latent became more coherent.
An independent review predicted this failure mode before the experiment. We
therefore treat writable pondering as a falsifiable coupling, not as an assumed
benefit.

Our main result separates optimization speed from attainable performance. At a
fixed budget of 240 steps, read-only pondering leads coupled pondering by 13.54
percentage points in held-out recall, a paired difference that excludes zero. At
960 steps, both arms reach the task ceiling. These observations support a
learning-speed penalty under the measured budget. They do not show that the
coupled architecture has lower capability, and they do not resolve behavior at
convergence on a task that remains below ceiling.

This paper makes five verifiable contributions:
\begin{enumerate}[leftmargin=*,itemsep=0.2em]
  \item It specifies a four-tier memory interface with explicit ownership,
  lifetime, and write boundaries.
  \item It defines a vertical variable-depth pondering process whose fast-tier
  write policy can be switched without changing the surrounding model.
  \item It reports a matched 12-seed ablation at 240 steps and an 8-seed
  longer-training control at 960 steps.
  \item It reports the negative-shaped positive that the predicted collapse of
  the energy margin did not occur in either measured setting.
  \item It provides an implementation with 5,920 lines in \texttt{memoir/},
  2,061 lines of tests, 53 Python files, and 70 passing tests, together with
  device-level timing and memory measurements.
\end{enumerate}

\section{Related Work}

\paragraph{Test-time memory and online weight updates.}
Linear attention can be written as a fast-weight program, which connects
sequence models to explicit online matrix updates \cite{schlag2021linear}.
Test-Time Training models replace a fixed recurrent state with a learned model
that updates while processing a sequence \cite{sun2024ttt}. Memorizing
Transformers retain selected key and value representations across context
segments \cite{wu2022memorizing}. Titans learns a neural long-term memory at
test time, while ATLAS studies how to optimize the memorization objective and
memory behavior \cite{behrouz2025titans,behrouz2025atlas}. End-to-End
Test-Time Training extends the training signal across long-context adaptation
\cite{tandon2025ttte2e}. The MIRAS framework relates test-time memorization,
attention, retention, and online optimization through a shared view
\cite{behrouz2025miras}. Memoir differs in the question it isolates: should a
fast tier be written inside each latent pondering iteration, or only between
forward passes?

\paragraph{Nested and continual learning.}
Nested Learning frames a model as interacting optimization problems operating
at different update frequencies \cite{behrouz2025nested}. Memoir implements a
concrete four-tier version of that view with per-sample fast and episodic
state, shared trainable slow parameters, and shared frozen parameters.
Elastic Weight Consolidation protects parameters associated with earlier
tasks by penalizing selected changes \cite{kirkpatrick2016ewc}. Our current
experiment does not measure long-horizon forgetting, so it cannot validate the
slow-tier consolidation story.

\paragraph{Adaptive and latent computation.}
Adaptive Computation Time learns how many recurrent steps to allocate
\cite{graves2016act}, and Universal Transformers apply a shared transition
recurrently across depth \cite{dehghani2018universal}. Recent latent-reasoning
work trains recurrent or continuous hidden computation before emitting text
\cite{geiping2025huginn,hao2024coconut}. Memoir combines adaptive recurrence
with writable weight state. This combination changes the recurrence operator
during the computation and motivates our direct ablation.

\paragraph{Energy and representation objectives.}
Energy-Based Transformers refine representations under an energy objective
\cite{gladstone2025ebt}. Bootstrap Your Own Latent and VICReg show two ways to
learn predictive representations while controlling collapse without negative
labels \cite{grill2020byol,bardes2021vicreg}. Memoir uses a future-latent
target and reports the coherent-versus-corrupted energy margin. We do not claim
that this small experiment establishes a general verifier.

\paragraph{Evaluation context.}
LongMemEval shows that access to long interaction histories does not guarantee
reliable use of past information \cite{wu2024longmemeval}. Test-time training
can also support few-shot adaptation beyond long-context language modeling
\cite{akyurek2024tttarc}. Analyses of hybrid linear attention map tradeoffs
among recurrent and attention mechanisms \cite{wang2025hybrid}. Work on why
language models hallucinate further cautions against treating fluent output as
evidence of calibrated knowledge \cite{kalai2025hallucinate}. These results
motivate direct memory, reasoning, and energy diagnostics rather than a single
aggregate score. Table~\ref{tab:nearest} locates Memoir relative to the closest
combinations in this literature.

\begin{table}[t]
\caption{Nearest systems compared across the four elements joined by Memoir.
Checks indicate an explicit mechanism, not identical semantics.}
\label{tab:nearest}
\centering
\small
\begin{tabular}{lcccc}
\toprule
System & Weights & Depth & Energy & Tiers \\
\midrule
Memoir & \cmark & \cmark & \cmark & \cmark \\
Nested Learning \cite{behrouz2025nested} & \cmark & \xmark & \xmark & \cmark \\
Titans \cite{behrouz2025titans} & \cmark & \xmark & \xmark & \xmark \\
TTT \cite{sun2024ttt} & \cmark & \xmark & \xmark & \xmark \\
Universal Transformer \cite{dehghani2018universal} & \xmark & \cmark & \xmark & \xmark \\
Energy-Based Transformer \cite{gladstone2025ebt} & \xmark & \cmark & \cmark & \xmark \\
CoCONUT \cite{hao2024coconut} & \xmark & \xmark & \xmark & \xmark \\
\bottomrule
\end{tabular}
\end{table}

\section{Method}

\subsection{Four weight tiers}

\begin{figure}[t]
\centering
\resizebox{\textwidth}{!}{%
\begin{tikzpicture}[
  tier/.style={draw, rounded corners, align=center, minimum width=2.7cm,
    minimum height=1.15cm, font=\small},
  lbl/.style={font=\footnotesize, gray!70, align=center}
]
  \node[tier, fill=green!12, draw=green!55!black] (w0) at (0,0)
    {$W_0$ fast\\\footnotesize written every forward pass};
  \node[tier, fill=blue!9, draw=blue!55, right=0.5cm of w0] (w1)
    {$W_1$ episodic\\\footnotesize gated write per segment};
  \node[tier, fill=orange!14, draw=orange!75!black, right=0.5cm of w1] (w2)
    {$W_2$ slow\\\footnotesize rare consolidation};
  \node[tier, fill=gray!10, draw=gray!55, densely dashed, right=0.5cm of w2] (w3)
    {$W_3$ frozen\\\footnotesize never updated, anchor};

  \draw[-{Latex[length=2.4mm]}, thick] (-1.6,-1.15) -- (13.2,-1.15);
  \node[lbl, anchor=north] at (0,-1.25) {within pass};
  \node[lbl, anchor=north] at (3.2,-1.25) {per session};
  \node[lbl, anchor=north] at (6.4,-1.25) {shared, durable};
  \node[lbl, anchor=north] at (9.6,-1.25) {immutable};
  \node[font=\footnotesize, anchor=north west] at (-1.6,-1.85)
    {increasing update interval};

  \draw[-{Latex[length=2mm]}, orange!70, thick]
    (w0.north) to[out=60,in=120] node[above,font=\footnotesize,black]
    {sleep: distill $W_0,W_1$ into $W_2$, reset fast} (w2.north);
\end{tikzpicture}%
}
\caption{The tier lifecycle. Information enters the per-sample fast tier, may
persist in the episodic tier, and reaches the shared slow tier only through the
proposed sleep consolidation, which then resets fast state. The frozen tier is
never written. Consolidation is a design component of the interface; this paper
does not evaluate it.}
\label{fig:timescales}
\end{figure}
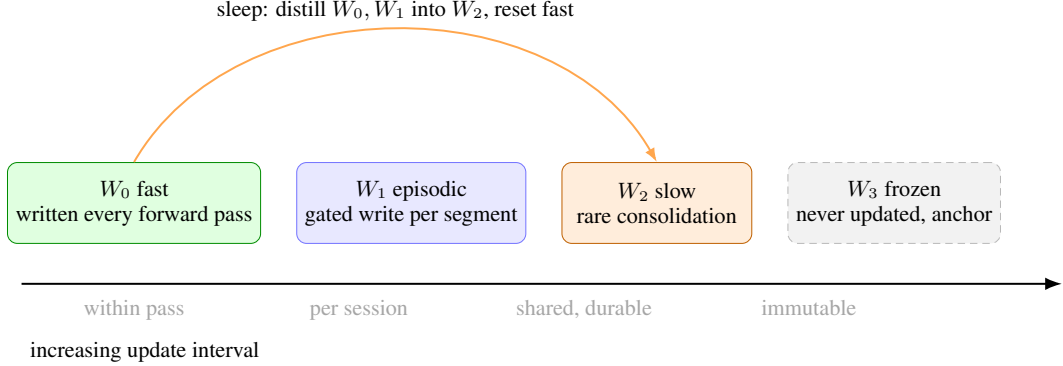

Let $B$ be batch size, $T$ sequence length, $H$ the number of heads, and $D_h$
the head dimension. The plastic state contains two per-sample matrices,
\begin{equation}
  W_0,W_1 \in \mathbb{R}^{B\times H\times D_h\times D_h}.
  \label{eq:plastic-shape}
\end{equation}
The caller carries these activations between segments and may detach them at a
credit boundary. $W_0$ is the fast tier and can change on every forward pass.
$W_1$ is the episodic tier and uses a smaller learned gate. The model owns two
shared parameter tiers. $W_2$ is trainable slow state, while $W_3$ has gradients
disabled and remains fixed. Equation~\ref{eq:plastic-shape} makes the key
ownership distinction explicit: $W_0$ and $W_1$ scale with the local batch,
whereas $W_2$ and $W_3$ do not.
Figure~\ref{fig:timescales} summarizes the lifecycle and write boundaries of
the four tiers.

The baseline write rule processes key, value, and write-strength tuples in
sequence. For one position, it applies
\begin{equation}
  W^{(t+1)} = W^{(t)} + \beta_t
  \left(v_t-W^{(t)}k_t\right)k_t^{\top}.
  \label{eq:delta}
\end{equation}
This delta update corrects the current value prediction and follows the
fast-weight interpretation of linear attention \cite{schlag2021linear}.
Memoir applies the same typed update-rule interface to either a hand-written
rule or a learned replacement. The experiments in this paper use the matched
architectural arms described in Section~\ref{sec:evaluation}; they do not test
a learned update rule.

\begin{figure}[t]
\centering
\begin{tikzpicture}[
  panel/.style={rounded corners=4pt, fill=gray!4, draw=gray!30, line width=0.6pt},
  tier/.style={rounded corners=2.5pt, draw, align=center, minimum width=3.45cm,
    minimum height=0.9cm, font=\small},
  nd/.style={rounded corners=2.5pt, draw=gray!60, fill=white, align=center,
    minimum height=0.72cm, font=\small},
  ttl/.style={font=\small\bfseries, anchor=west},
  sub/.style={font=\scriptsize, gray!80!black, anchor=west},
  lab/.style={font=\scriptsize, gray!60!black},
  fwd/.style={-{Latex[length=2mm]}, gray!70!black, line width=0.75pt},
  wr/.style={-{Latex[length=2.4mm]}, red!75!black, line width=1.5pt,
    dash pattern=on 5pt off 2pt},
  sg/.style={-{Latex[length=1.8mm]}, densely dotted, violet!70!black, line width=0.8pt},
]

\draw[panel] (0,-0.10) rectangle (4.30,6.55);
\draw[panel] (4.60,-0.10) rectangle (9.20,6.55);
\draw[panel] (9.50,-0.10) rectangle (13.60,6.55);

\node[ttl] at (0.22,6.28) {MEMORY};
\node[sub] at (0.22,6.00) {four timescales};
\node[ttl] at (4.82,6.28) {PONDER};
\node[sub] at (4.82,6.00) {variable depth, learned halt};
\node[ttl] at (9.72,6.28) {PREDICT $+$ SCORE};
\node[sub] at (9.72,6.00) {future latent, energy readout};

\node[tier, fill=green!12, draw=green!55!black] (w0) at (2.35,4.62)
  {$W_0$ fast\\[-2pt]\scriptsize written every forward pass};
\node[tier, fill=blue!9,  draw=blue!55]        (w1) at (2.35,3.52)
  {$W_1$ episodic\\[-2pt]\scriptsize gated write per segment};
\node[tier, fill=orange!14, draw=orange!75!black] (w2) at (2.35,2.42)
  {$W_2$ slow\\[-2pt]\scriptsize rare consolidation};
\node[tier, fill=gray!8, draw=gray!55, densely dashed] (w3) at (2.35,1.32)
  {$W_3$ frozen\\[-2pt]\scriptsize never updated, anchor};

\draw[-{Latex[length=1.8mm]}, gray!55] (0.42,5.07) -- (0.42,0.90);
\node[lab, rotate=90, anchor=south] at (0.28,2.98) {increasing update interval};

\node[lab, anchor=west] at (0.55,0.55)
  {$W_0,W_1$: per-sample, $B{\times}H{\times}D_h{\times}D_h$};
\node[lab, anchor=west] at (0.55,0.24)
  {sleep: distill $W_0,W_1$ into $W_2$, reset fast};

\draw[rounded corners=4pt, draw=gray!50, line width=0.7pt]
  (4.85,2.05) rectangle (8.95,4.95);
\node[lab] at (6.50,4.70) {$\ell = 1,\dots,L$, per-position halt};

\node[nd, minimum width=1.05cm] (hl) at (5.62,3.55) {$h^{(\ell)}$};
\node[nd, minimum width=1.25cm, fill=cyan!8, draw=cyan!60!black] (fb) at (6.90,3.55)
  {$F_\theta$};
\node[nd, minimum width=1.15cm] (hn) at (8.20,3.55) {$h^{(\ell+1)}$};
\draw[fwd] (hl) -- (fb);
\draw[fwd] (fb) -- (hn);
\draw[fwd] (hn.south) -- ++(0,-0.62) -| (hl.south);
\node[lab] at (6.90,2.30) {halting head: continue or emit};

\draw[gray!55, line width=0.6pt] (w0.east) -- (4.72,3.42);
\draw[gray!55, line width=0.6pt] (w1.east) -- (4.72,3.42);
\draw[gray!55, line width=0.6pt] (w2.east) -- (4.72,3.42);
\draw[gray!55, line width=0.6pt] (w3.east) -- (4.72,3.42);
\fill[gray!55] (4.72,3.42) circle (1.4pt);
\draw[fwd] (4.72,3.42) -- (hl.west |- 0,3.42);
\node[lab, anchor=south] at (4.88,3.55) {read};

\node[nd, fill=red!5, draw=red!50, minimum width=3.4cm, inner ysep=2.5pt,
  font=\scriptsize] (rule) at (6.35,5.32)
  {update rule $(W,k,v,\beta)\to W'$\\[-1pt]\tiny delta (fixed) or RuleNet (learned)};
\draw[wr] (hn.north) |- (rule.east);
\draw[wr] (rule.west) -| (w0.north);
\node[font=\scriptsize\bfseries, red!75!black, anchor=south, align=center]
  at (3.15,5.46) {write every iteration\\[-2pt]\scriptsize\itshape the tested coupling};

\node[nd, minimum width=1.55cm] (ctx) at (10.45,4.35) {context $h$};
\node[nd, minimum width=1.35cm, fill=violet!7, draw=violet!55] (pred) at (12.55,4.35)
  {$P_\psi$\\[-2pt]\scriptsize predictor};
\node[nd, minimum width=1.35cm] (zhat) at (12.55,3.10) {$\hat z_{t+\Delta}$};
\node[nd, minimum width=1.85cm, densely dotted, draw=violet!55] (teach) at (10.45,3.10)
  {EMA teacher\\[-2pt]\scriptsize $\bar z_{t+\Delta}$, stop-grad};
\node[nd, minimum width=1.85cm, fill=violet!12, draw=violet!65!black] (en) at (11.50,1.80)
  {$E_\phi(\hat z,\bar z)$\\[-2pt]\scriptsize energy};

\draw[fwd] (hn.east) -- ++(0.28,0) |- (ctx.west);
\draw[fwd] (ctx) -- (pred);
\draw[fwd] (pred) -- (zhat);
\draw[fwd] (zhat.south) -- ++(0,-0.35) -| ($(en.north)+(0.45,0)$);
\draw[sg]  (teach.south) -- ++(0,-0.35) -| ($(en.north)+(-0.45,0)$);
\node[lab, anchor=north, align=center] at (11.50,1.28)
  {$e_t$ low $=$ coherent\\high: ponder more or abstain};
\draw[sg] (en.west) -- ++(-0.55,0) -- (8.95,2.60);
\node[font=\scriptsize, violet!70!black, anchor=east] at (10.50,1.56) {informs halting};

\draw[fwd] (0.25,-0.62) -- ++(0.55,0);
\node[lab, anchor=west] at (0.90,-0.62) {forward read};
\draw[wr] (3.05,-0.62) -- ++(0.55,0);
\node[lab, anchor=west] at (3.70,-0.62) {tested write (hypothesis)};
\draw[sg] (6.95,-0.62) -- ++(0.55,0);
\node[lab, anchor=west] at (7.60,-0.62) {stop-grad / EMA};
\draw[densely dashed, gray!55, line width=0.8pt] (10.05,-0.62) -- ++(0.55,0);
\node[lab, anchor=west] at (10.70,-0.62) {frozen};

\end{tikzpicture}
\caption{One Memoir layer. Left: four weight tiers ordered by update interval,
with per-sample fast and episodic state and shared slow and frozen parameters.
Middle: the variable-depth pondering loop; the highlighted red path writes the
fast tier through the update rule on every active iteration and is the coupling
tested in Section 5. Right: future-latent prediction against a stop-gradient
EMA teacher, scored by an energy head that informs halting.}
\label{fig:architecture}
\end{figure}
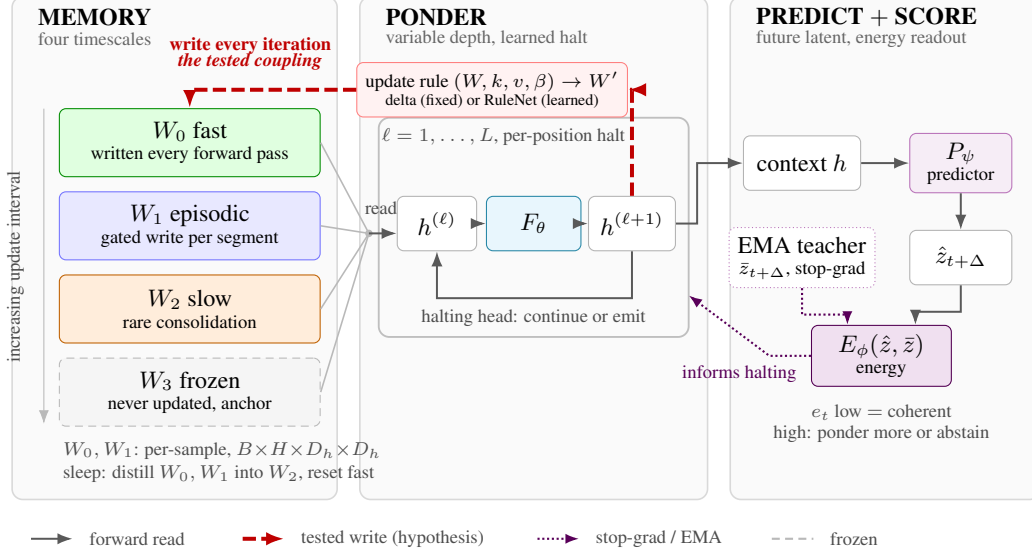

\subsection{Variable-depth pondering and tier writes}

Figure~\ref{fig:architecture} shows the data path. At recurrent step $\ell$,
the shared block reads the current latent and available tiers, then proposes
\begin{equation}
  h^{(\ell+1)} = F_{\theta}\!\left(h^{(\ell)},W_0^{(\ell)},W_1^{(\ell)},W_2,W_3\right).
  \label{eq:ponder}
\end{equation}
A halting head emits a probability for each position, following the adaptive
computation pattern of recurrent models \cite{graves2016act}. Active positions
continue until the threshold or configured cap is reached. In the coupled arm,
the same active mask gates tier writes after each recurrent transition. In the
read-only arm, pondering does not write $W_0$ or $W_1$; the surrounding
forward-pass boundary retains the normal update path. Equation~\ref{eq:ponder}
therefore describes a stationary operator only for the read-only arm.

\begin{algorithm}[t]
\caption{Pondering with optional tier writes}
\label{alg:ponder}
\begin{algorithmic}[1]
\REQUIRE latent $h$, tiers $(W_0,W_1)$, maximum steps $L$, coupled flag $c$
\STATE initialize active mask $a$ to true and accumulated halt mass to zero
\FOR{$\ell=1$ to $L$}
  \STATE read $(W_0,W_1)$ for active positions and compute candidate $\tilde h$
  \STATE set $h \leftarrow \tilde h$ where $a$ is true
  \STATE compute halt probabilities and update $a$
  \IF{$c$ is true}
    \STATE derive keys, values, and write strengths from $h$
    \STATE update $W_0$ by Equation~\ref{eq:delta} where $a$ was active
    \STATE update $W_1$ through its gated write path
  \ENDIF
  \IF{all positions halted}
    \STATE break
  \ENDIF
\ENDFOR
\RETURN $h$, $(W_0,W_1)$, halt steps, ponder cost
\end{algorithmic}
\end{algorithm}

Algorithm~\ref{alg:ponder} exposes the tested coupling. It also exposes why an
energy signal might fail: the latent and its memory-conditioned scoring context
can change in the same loop. The ablation preserves parameter counts and the
declared forward compute match while removing only those within-loop writes.

\subsection{Future-latent and energy objectives}

Memoir predicts a future target latent from the current representation. A
teacher encoder supplies stop-gradient targets, and an exponential moving
average updates the teacher. This design follows predictive representation
learning rather than token reconstruction \cite{grill2020byol,bardes2021vicreg}.
For prediction horizon $\Delta$, the model minimizes a latent regression loss
between $\hat z_{t+\Delta}$ and $z_{t+\Delta}$. An energy head also scores
coherent and corrupted pairs. We report the margin
\begin{equation}
  m = \mathbb{E}[E(\text{corrupted})]-\mathbb{E}[E(\text{coherent})],
  \label{eq:margin}
\end{equation}
so a positive value means coherent pairs receive lower energy. The independent
failure prediction expected within-loop memory rewriting to make this head
gameable. Section~\ref{sec:evaluation} reports that the measured margin in
Equation~\ref{eq:margin} instead grew during training.

\section{Implementation}

The implementation contains 5,920 lines under \texttt{memoir/}, 2,061 lines of
tests, and 53 Python files. The current test run contains 70 passing tests.
The frozen interface contract fixes tensor shapes, ownership, and return types
for update rules, pondering, energy, and latent prediction. It defines $W_0$
and $W_1$ as per-sample activations with shape
$B\times H\times D_h\times D_h$. It defines $W_2$ as a shared trainable
parameter and $W_3$ as a shared parameter with gradients disabled. A tier-state
detach operation cuts graphs at caller-selected segment boundaries.

The optimized delta rule preserves the sequential recurrence in
Equation~\ref{eq:delta}. Within each token chunk, it constructs the strictly
lower triangular key Gram matrix and solves a unit lower triangular system.
Python iterates over chunks rather than token positions. The pondering hot path
uses an update-only entry point that avoids materializing diagnostics discarded
by training. Plastic state stays in fp32 on the measured system. Explicit
batched matrix multiplications serve tier reads, and activation checkpointing
is available per pondering iteration. Section~\ref{sec:evaluation} reports the
measured effect of these changes.

\section{Experimental Evaluation}
\label{sec:evaluation}

\subsection{Associative recall protocol}

The task is procedural associative recall with key interference. It uses 64
possible values, 16 distractors, and interference 0.85. Arm A couples pondering
and memory: every pondering iteration writes the fast tier $W_0$. Arm B reads
$W_0$ during pondering but writes it only between forward passes. Both arms
have 81,738 total parameters and 76,362 trainable parameters. The experiment
asserts matching declared forward multiply-accumulate counts in code. It also
holds the optimizer, schedule, data, and seed set fixed.

We report the mean held-out recall and the 95 percent interval across seeds. The
primary budget uses 240 training steps and 12 seeds. The longer-training control
uses 960 steps and 8 seeds. The two arms are exactly paired: for each seed they
start from bitwise identical weights and consume the same batch at every step,
so the per-seed difference is the appropriate statistic and we report a paired
test on it. The per-arm intervals in Table~\ref{tab:recall-240} use a normal
quantile and are descriptive only; we do not rest the claim on whether they
separate, because that separation depends on the quantile chosen.

\begin{table}[t]
\caption{Held-out recall after 240 training steps across 12 seeds. The final
column is Arm B minus Arm A.}
\label{tab:recall-240}
\centering
\small
\begin{tabular}{lccc}
\toprule
Arm & Mean & 95 percent interval & Difference \\
\midrule
A: coupled & 0.5203 & [0.4522, 0.5883] & \\
B: read-only & 0.6557 & [0.5953, 0.7160] & +0.1354 \\
\bottomrule
\end{tabular}
\end{table}

\begin{figure}[t]
\centering
\begin{tikzpicture}[x=1cm,y=4.4cm]
  \def\ymin{0.00}\def\ymax{0.80}
  \foreach \v in {0.0,0.1,0.2,0.3,0.4,0.5,0.6,0.7}{
    \draw[gray!22] (0,\v) -- (6.4,\v);
    \node[left,font=\footnotesize,gray!60] at (0,\v) {\v};
  }
  \draw[-{Latex[length=2mm]}] (0,\ymin) -- (0,\ymax);
  \node[rotate=90,anchor=south,font=\small] at (-0.95,0.40) {held-out recall};

  \def\amx{1.7}
  \fill[red!12] (\amx-0.6,0) rectangle (\amx+0.6,0.5203);
  \draw[red!55,line width=0.9pt] (\amx-0.6,0) rectangle (\amx+0.6,0.5203);
  \draw[red!70,line width=1.1pt] (\amx,0.4522) -- (\amx,0.5883);
  \draw[red!70,line width=1.1pt] (\amx-0.18,0.4522) -- (\amx+0.18,0.4522);
  \draw[red!70,line width=1.1pt] (\amx-0.18,0.5883) -- (\amx+0.18,0.5883);
  \node[font=\footnotesize\bfseries] at (\amx,0.63) {0.5203};
  \node[font=\footnotesize,align=center,anchor=south] at (\amx,0.03) {Arm A\\coupled};

  \def\bmx{4.7}
  \fill[teal!10] (\bmx-0.6,0) rectangle (\bmx+0.6,0.6557);
  \draw[teal!60!black,line width=0.9pt] (\bmx-0.6,0) rectangle (\bmx+0.6,0.6557);
  \draw[teal!70!black,line width=1.1pt] (\bmx,0.5953) -- (\bmx,0.7160);
  \draw[teal!70!black,line width=1.1pt] (\bmx-0.18,0.5953) -- (\bmx+0.18,0.5953);
  \draw[teal!70!black,line width=1.1pt] (\bmx-0.18,0.7160) -- (\bmx+0.18,0.7160);
  \node[font=\footnotesize\bfseries] at (\bmx,0.755) {0.6557};
  \node[font=\footnotesize,align=center,anchor=south] at (\bmx,0.03) {Arm B\\read-only};

  \draw[gray!45,dashed] (\amx+0.6,0.5203) -- (3.2,0.5203);
  \draw[gray!45,dashed] (3.2,0.6557) -- (\bmx-0.6,0.6557);
  \draw[{Latex[length=1.5mm]}-{Latex[length=1.5mm]},gray!70,line width=0.9pt]
    (3.2,0.5203) -- (3.2,0.6557);
  \node[font=\small\bfseries,anchor=south,align=center] at (3.0,0.665)
    {$+0.1354$};
  \node[font=\scriptsize,align=center,fill=white,inner sep=1.5pt] at (3.2,0.28)
    {paired $t=3.23$\\10 of 12 seeds};
\end{tikzpicture}
\caption{Held-out recall at the 240 step budget across 12 seeds. Bars are means,
whiskers are the reported normal-quantile intervals, shown as descriptive
context. The paired per-seed test in the text is the primary statistic.}
\label{fig:result}
\end{figure}
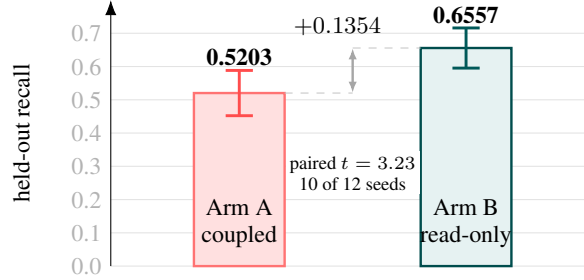

Table~\ref{tab:recall-240} shows a 0.1354 advantage for read-only pondering at
the fixed 240-step budget. Figure~\ref{fig:result} shows the same means and
descriptive intervals. Because the arms are paired per seed, we test the 12
per-seed differences directly. The mean difference is 0.1354 with a standard
deviation of 0.1453, giving a paired $t$ of 3.23 on 11 degrees of freedom and a
95 percent interval on the difference of $[0.0431, 0.2277]$, which excludes
zero. Read-only pondering wins on 10 of the 12 seeds. We interpret this as a
measured learning-speed penalty from within-loop writes under this budget. We do
not interpret it as a capability penalty.

\begin{figure}[t]
\centering
\begin{tikzpicture}[x=0.030cm,y=85cm]
  \def\ymin{-0.002}\def\ymax{0.026}
  \foreach \v in {0.000,0.005,0.010,0.015,0.020,0.025}{
    \draw[gray!22] (0,\v) -- (240,\v);
    \node[left,font=\footnotesize,gray!60] at (0,\v) {\v};
  }
  \foreach \s in {0,40,80,120,160,200,240}{
    \node[below,font=\footnotesize,gray!60] at (\s,\ymin) {\s};
  }
  \draw[-{Latex[length=2mm]}] (0,\ymin) -- (0,\ymax);
  \draw[-{Latex[length=2mm]}] (0,\ymin) -- (250,\ymin);
  \node[below,font=\small] at (125,-0.0078) {training step};
  \node[rotate=90,anchor=south,font=\small] at (-30,0.012) {energy margin};

  \draw[gray!50,dashed] (0,0) -- (240,0);

  \draw[teal!70,line width=1.3pt]
    (0,-0.0001) -- (40,0.0035) -- (80,0.0037) -- (120,0.0100)
    -- (160,0.0181) -- (200,0.0224) -- (240,0.0231);
  \foreach \s/\v in {0/-0.0001,40/0.0035,80/0.0037,120/0.0100,160/0.0181,200/0.0224,240/0.0231}{
    \fill[teal!80] (\s,\v) circle (2.6pt);
  }
  \node[anchor=west,font=\footnotesize\bfseries,teal!70] at (150,0.0135)
    {margin grows, no collapse};
  \node[anchor=east,font=\footnotesize] at (238,0.0247) {0.0231};
\end{tikzpicture}
\caption{The held-out energy margin grew and held during the twelve seed run.
The predicted collapse, in which writes lower energy by making memory agree
with the current latent, did not occur.}
\label{fig:energy}
\end{figure}
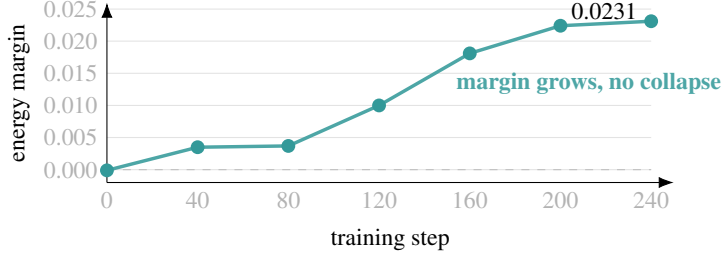

The coherent-versus-corrupted energy margin over training was
$-0.0001$, $0.0035$, $0.0037$, $0.0100$, $0.0181$, $0.0224$, and $0.0231$.
Thus the margin became positive and grew rather than collapsing. The
halt-at-cap frequency was 11.04 percent. The predicted route to failure, in
which coupled writes corrupt the energy head by letting memory move toward the
current latent, did not appear in these diagnostics. Figure~\ref{fig:energy}
traces the measured margin across training.

\begin{table}[t]
\caption{Held-out recall after 960 training steps across 8 seeds. Both arms
reach the ceiling, so the control cannot measure a nonzero convergence gap.}
\label{tab:recall-960}
\centering
\small
\begin{tabular}{lccc}
\toprule
Arm & Mean & 95 percent interval & Difference \\
\midrule
A: coupled & 1.0000 & [1.0000, 1.0000] & \\
B: read-only & 1.0000 & [1.0000, 1.0000] & 0.0000 \\
\bottomrule
\end{tabular}
\end{table}

Table~\ref{tab:recall-960} changes the interpretation of the short-budget
result. After 960 steps, both arms reach 1.0000 and the difference is exactly
zero. The energy margin ends at 0.1353. This control shows that the 240-step gap
is not permanent on this task. It does not establish equality at convergence
on a non-saturating task because the control itself is ceiling limited. That
question remains unmeasured.

\subsection{Discarded saturated task}

An earlier run, described in the study brief as using 8 values and 5
distractors, saturated both arms at 1.0000. The archived summary records 7
distractors for the same run. The analysis program still printed a confident
falsified verdict. We discarded that run as uninformative and increased task
difficulty. This failure of the evaluation setup, rather than the model,
motivated the 64-value, 16-distractor protocol and reporting raw ceiling
behavior next to any automated verdict.

\subsection{Engineering measurements}

We measured engineering paths on an NVIDIA GeForce RTX 5090 Laptop GPU with
24 GB of memory and compute capability 12.0, using PyTorch 2.11.0+cu128. The
timings use CUDA events, warmup, and 50 measured iterations. Table
\ref{tab:engineering} reports only measured quantities supplied by the
benchmark records. Effective bandwidth is reported for the delta-rule path.

\begin{table}[t]
\caption{Engineering measurements. Memory values are available for the
delta-rule forward path only.}
\label{tab:engineering}
\centering
\scriptsize
\begin{tabular}{lccccc}
\toprule
Path & Before ms & After ms & Speedup & Before MiB & After MiB \\
\midrule
Delta-rule forward & 0.9073 & 0.3509 & 2.59$\times$ & 40.3 & 34.2 \\
Pondering forward & 14.07 & 9.09 & 1.55$\times$ & n/a & n/a \\
Full train step & 90.02 & 86.11 & 1.05$\times$ & n/a & n/a \\
\bottomrule
\end{tabular}
\end{table}

For the delta-rule forward path, effective bandwidth increased from 18.6 GB/s
to 48.0 GB/s while peak memory fell from 40.3 MiB to 34.2 MiB. The 2.59$\times$
kernel-level improvement becomes a 1.55$\times$ pondering improvement and a
1.05$\times$ full-step improvement. This narrowing identifies the remaining
work outside the isolated update rule without projecting an unmeasured speedup.

\section{Limitations}

The most important limitation is that our ablation does not isolate the variable
its name suggests. Arm A writes the fast tier once per pondering iteration and
again between forward passes, so it performs roughly seven times as many writes
per episode as Arm B, and the in-loop writes do not decay. The task is built so
that key interference is the bottleneck, which is exactly the regime where
additional writes into the same key space are expected to hurt on their own.
The measurement therefore supports the statement that this coupled
configuration learns more slowly. It does not separate the timing of the writes
from their volume. The control that would separate them is a third arm applying
the same ponder-derived writes once between forward passes at matched write
magnitude, and we have not run it. Any mechanistic reading of the result should
wait for that arm.

A second asymmetry runs the other way. The projections that produce in-loop
writes exist in both arms for parameter matching, but in Arm B they receive no
gradient and do not affect its output, so roughly 11 percent of that arm's
trainable parameters are inert. This handicaps the arm that won, so it does not
threaten the direction of the result, but the parameter match is more exact than
the functional match.

The central result comes from one procedural recall family at one difficult
configuration. The 960-step control saturates, so performance at
convergence on a non-saturating task remains unmeasured. It is also a fresh run
with its own schedule rather than a continuation of the 240-step run. The
experiment does
not test natural language, long-horizon continual learning, consolidation into
$W_2$, resistance to catastrophic forgetting, or a learned update rule. The
energy margin is a task-specific diagnostic and does not establish calibrated
verification or hallucination detection. LongMemEval and analyses of
hallucination illustrate why broader memory and reliability evaluations remain
necessary \cite{wu2024longmemeval,kalai2025hallucinate}.

The parameter and declared forward-compute match controls important confounds,
but it does not make the two runtime traces identical. Within-loop writes can
change memory traffic and optimization dynamics by design. The engineering
measurements come from one GPU, one software build, and the stated benchmark
procedure. They should not be generalized to other hardware without new
measurements.

\section{Conclusion}

Memoir provides a concrete interface for joining fast and episodic weight
state, shared slow and frozen tiers, adaptive latent recurrence, and a
future-latent energy objective. Its defining coupling allows pondering to write
the same fast memory that it reads. On the measured associative-recall task,
that coupling slows learning at 240 steps: the read-only arm leads by 0.1354 on
a paired test over 12 seeds. The result does not support an outright
refutation, and it does not separate write timing from write volume. At 960
steps, both arms reach 1.0000 and the gap disappears. Since
that control is ceiling limited, any convergence difference on a
non-saturating task remains unknown. The predicted energy-head corruption also
did not occur; the margin grew and held. The next decisive experiment should
retain headroom through convergence while preserving the same matched arms and
diagnostics.


\small
\begin{thebibliography}{20}

\bibitem[Akyurek et al.(2024)]{akyurek2024tttarc}
Ekin Akyurek, Mehul Damani, Adam Zweiger, Linlu Qiu, Han Guo, Jyothish Pari,
Yoon Kim, and Jacob Andreas.
The Surprising Effectiveness of Test-Time Training for Few-Shot Learning.
arXiv:2411.07279, 2024.

\bibitem[Bardes et al.(2021)]{bardes2021vicreg}
Adrien Bardes, Jean Ponce, and Yann LeCun.
VICReg: Variance-Invariance-Covariance Regularization for Self-Supervised Learning.
arXiv:2105.04906, 2021.

\bibitem[Behrouz et al.(2025a)]{behrouz2025atlas}
Ali Behrouz, Zeman Li, Praneeth Kacham, Majid Daliri, Yuan Deng, Peilin Zhong,
Meisam Razaviyayn, and Vahab Mirrokni.
ATLAS: Learning to Optimally Memorize the Context at Test Time.
arXiv:2505.23735, 2025.

\bibitem[Behrouz et al.(2025b)]{behrouz2025miras}
Ali Behrouz, Meisam Razaviyayn, Peilin Zhong, and Vahab Mirrokni.
It Is All Connected: A Journey Through Test-Time Memorization, Attentional
Bias, Retention, and Online Optimization.
arXiv:2504.13173, 2025.

\bibitem[Behrouz et al.(2025c)]{behrouz2025nested}
Ali Behrouz, Meisam Razaviyayn, Peilin Zhong, and Vahab Mirrokni.
Nested Learning: The Illusion of Deep Learning Architectures.
arXiv:2512.24695, 2025.

\bibitem[Behrouz et al.(2025d)]{behrouz2025titans}
Ali Behrouz, Peilin Zhong, and Vahab Mirrokni.
Titans: Learning to Memorize at Test Time.
arXiv:2501.00663, 2025.

\bibitem[Dehghani et al.(2018)]{dehghani2018universal}
Mostafa Dehghani, Stephan Gouws, Oriol Vinyals, Jakob Uszkoreit, and Lukasz Kaiser.
Universal Transformers.
arXiv:1807.03819, 2018.

\bibitem[Geiping et al.(2025)]{geiping2025huginn}
Jonas Geiping, Sean McLeish, Neel Jain, John Kirchenbauer, Siddharth Singh,
Brian R. Bartoldson, Bhavya Kailkhura, Abhinav Bhatele, and Tom Goldstein.
Scaling up Test-Time Compute with Latent Reasoning: A Recurrent Depth Approach.
arXiv:2502.05171, 2025.

\bibitem[Gladstone et al.(2025)]{gladstone2025ebt}
Alexi Gladstone, Ganesh Nanduru, Md Mofijul Islam, Peixuan Han, Hyeonjeong Ha,
Aman Chadha, Yilun Du, Heng Ji, Jundong Li, and Tariq Iqbal.
Energy-Based Transformers Are Scalable Learners and Thinkers.
arXiv:2507.02092, 2025.

\bibitem[Graves(2016)]{graves2016act}
Alex Graves.
Adaptive Computation Time for Recurrent Neural Networks.
arXiv:1603.08983, 2016.

\bibitem[Grill et al.(2020)]{grill2020byol}
Jean-Bastien Grill, Florian Strub, Florent Altche, et al.
Bootstrap Your Own Latent: A New Approach to Self-Supervised Learning.
arXiv:2006.07733, 2020.

\bibitem[Hao et al.(2024)]{hao2024coconut}
Shibo Hao, Sainbayar Sukhbaatar, DiJia Su, Xian Li, Zhiting Hu, Jason Weston,
and Yuandong Tian.
Training Large Language Models to Reason in a Continuous Latent Space.
arXiv:2412.06769, 2024.

\bibitem[Kalai et al.(2025)]{kalai2025hallucinate}
Adam Tauman Kalai, Ofir Nachum, Santosh S. Vempala, and Edwin Zhang.
Why Language Models Hallucinate.
arXiv:2509.04664, 2025.

\bibitem[Kirkpatrick et al.(2016)]{kirkpatrick2016ewc}
James Kirkpatrick, Razvan Pascanu, Neil Rabinowitz, et al.
Overcoming Catastrophic Forgetting in Neural Networks.
arXiv:1612.00796, 2016.

\bibitem[Schlag et al.(2021)]{schlag2021linear}
Imanol Schlag, Kazuki Irie, and Juergen Schmidhuber.
Linear Transformers Are Secretly Fast Weight Programmers.
arXiv:2102.11174, 2021.

\bibitem[Sun et al.(2024)]{sun2024ttt}
Yu Sun, Xinhao Li, Karan Dalal, Jiarui Xu, Arjun Vikram, Genghan Zhang,
Yann Dubois, Xinlei Chen, Xiaolong Wang, Sanmi Koyejo, Tatsunori Hashimoto,
and Carlos Guestrin.
Learning to (Learn at Test Time): RNNs with Expressive Hidden States.
arXiv:2407.04620, 2024.

\bibitem[Tandon et al.(2025)]{tandon2025ttte2e}
Arnuv Tandon, Karan Dalal, Xinhao Li, et al.
End-to-End Test-Time Training for Long Context.
arXiv:2512.23675, 2025.

\bibitem[Wang et al.(2025)]{wang2025hybrid}
Dustin Wang, Rui-Jie Zhu, Steven Abreu, et al.
A Systematic Analysis of Hybrid Linear Attention.
arXiv:2507.06457, 2025.

\bibitem[Wu et al.(2024)]{wu2024longmemeval}
Di Wu, Hongwei Wang, Wenhao Yu, Yuwei Zhang, Kai-Wei Chang, and Dong Yu.
LongMemEval: Benchmarking Chat Assistants on Long-Term Interactive Memory.
arXiv:2410.10813, 2024.

\bibitem[Wu et al.(2022)]{wu2022memorizing}
Yuhuai Wu, Markus N. Rabe, DeLesley Hutchins, and Christian Szegedy.
Memorizing Transformers.
arXiv:2203.08913, 2022.

\end{thebibliography}
\end{document}